\documentclass[twoside,11pt]{article}

\usepackage{jmlr2e}
\usepackage{blindtext}
\usepackage{amsmath, amssymb, bbm, amsfonts}
\usepackage{footnote}
\usepackage{comment}
\usepackage{float}
\usepackage{xcolor}
\usepackage{multirow}
\usepackage{graphicx}  
\usepackage{booktabs}
\usepackage{graphicx}
\usepackage{subcaption}
\usepackage{changepage}

\usepackage{hyperref, cleveref, placeins}
\newcommand{\ind}{\mathbbm{1}}
\newcommand{\argmin}{\operatornamewithlimits{arg\,min}}

\def \Ic{\mathcal I}
\def \Lc{\mathcal L}

\def \AF{\mathcal {AF}}

\def \R{\mathbb{R}}

\def \xx{\mathbf{x}}
\def \ff{\mathbf{f}}

\def \ww{\mathbf{w}}

\def \uu{\mathbf{u}}

\DeclareMathOperator*{\argmax}{arg\:max}

\usepackage{lastpage}
\jmlrheading{26}{2025}{Submitted}{; Revised }{}{21-0000}{Bertsimas and Cui.}

\ShortHeadings{Bertsimas and Cui}{Bertsimas and Cui}
\firstpageno{1}

\begin{document}

\title{Adaptive Forests For Classification}

\author{\name Dimitris Bertsimas \email dbertsim@mit.edu \\
       \addr Sloan School of Management and Operations Research Center\\
       Massachusetts Institute of Technology\\
       Cambridge, MA 02139, USA
       \AND
       \name Yubing Cui \email yubingc@mit.edu \\
       \addr Operations Research Center\\
       Massachusetts Institute of Technology\\
       Cambridge, MA 02139, USA}
       
\editor{}

\maketitle

\begin{abstract}

Random Forests (RF) and Extreme Gradient Boosting (XGBoost) are two of the most widely used and highly performing classification and regression models. They aggregate equally weighted CART trees, generated randomly in RF or sequentially in XGBoost. In this paper, we propose Adaptive Forests (AF), a novel approach that adaptively selects the weights of the underlying CART models. AF combines (a) the Optimal Predictive-Policy Trees (OP2T) framework to prescribe tailored, input-dependent unequal weights to trees and (b) Mixed Integer Optimization (MIO) to refine weight candidates dynamically, enhancing overall performance. We demonstrate that AF consistently outperforms RF, XGBoost, and other weighted RF in binary and multi-class classification problems over 20+ real-world datasets.

\end{abstract}

\begin{keywords}
  decision trees, adaptive learning, ensemble modeling,  optimization, weighted random forest 
\end{keywords}

\section{Introduction}

Since \cite{breiman1984cart} introduced Classification and Regression Trees (CART), decision trees have become fundamental algorithms in machine learning research and have been extensively applied in real-world problems. Moreover, decision trees have laid the foundation for the development of tree-based methods. Random Forests, introduced in \cite{breiman2001rf}, leverages ensembles of independent trees to enhance predictive performance through the wisdom of crowds, while gradient boosted trees employ sequential ensembles where each tree iteratively improves on its predecessor. XGBoost, introduced by \cite{chen2016xgboost}, is an optimized implementation of gradient boosted trees that has gained widespread recognition for its exceptional speed, scalability, and accuracy across diverse machine learning tasks.

Random Forests creates multiple independent CART models using different subsets of the data and aggregates their predictions through averaging for regression or majority vote for classification. Efforts to enhance this approach have involved weighting trees unequally based on their accuracy with Out-of-Bag (OOB) data, but this weighting is uniformly applied across all data points. The question arises whether we can improve this further by applying unequal weights to trees and adapting these weights dynamically for different data inputs.

Let us motivate with an example from human experience. Consider the decision-making process about whether to administer a particular treatment to a patient. A panel of doctors, each specializing in a different area, convene to discuss the case. In such scenarios, the opinions of these doctors are not typically weighted equally or simply averaged. It is neither reasonable nor optimal to consistently prioritize certain doctors’ opinions across all patients.   Instead, more weight may be assigned to certain doctors’ views based on various factors such as their prior interactions with the patient, the relevance of their specialties to the patient’s initial diagnosis, or the confidence in their judgments. The challenge then is to determine a personalized weighting for each doctor’s input that aligns with the patient’s unique characteristics and is more effective in improving outcomes than a uniform averaging approach.

Analogously, we propose Adaptive Forests, an algorithm that seeks to train diverse experts specialized in different areas, represented by base learner CARTs, and devises a weighting policy that assigns tailored weights based on input characteristics, enhancing the overall prediction accuracy. By utilizing a novel ensemble framework, OP2T \cite{OP2T}—detailed in Section \ref{OP2T}—that is both interpretable and adaptive, we assign input-specific, unequal weights to trees. This approach forms the basis of Adaptive Forests, a new tree-based model designed for general classification tasks that consistently outperforms Random Forests, XGBoost, and other weighted random forest algorithms. Given its strong performance, we hope Adaptive Forests will become a leading method in tree-based modeling. 

\begin{figure}[p]
    \begin{adjustwidth}{-2.2cm}{-2.1cm}  
    \centering
    \textbf{Human Decision Making} \\[5pt]
    
    \begin{minipage}{0.48\textwidth}  
        \centering
        \includegraphics[width=1.1\textwidth]{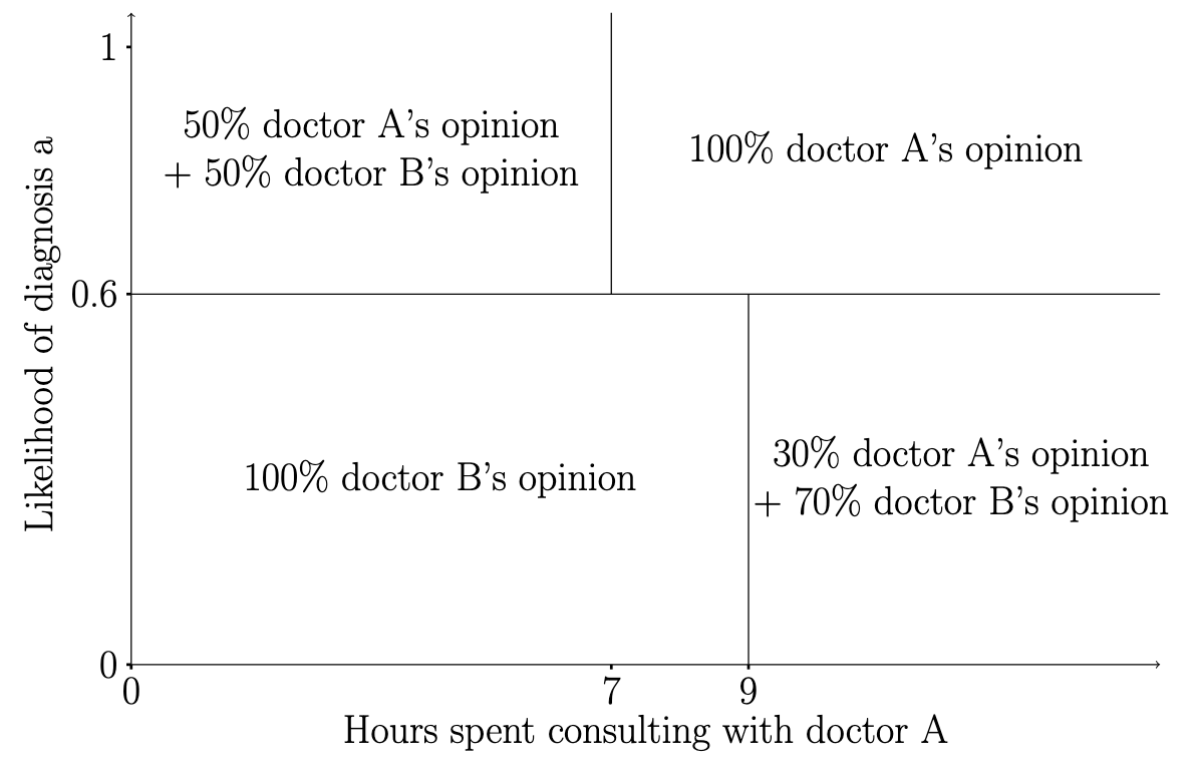}
    \end{minipage}
    \hfill
    \begin{minipage}{0.7\textwidth}  
        \centering
        \includegraphics[width=1\textwidth]{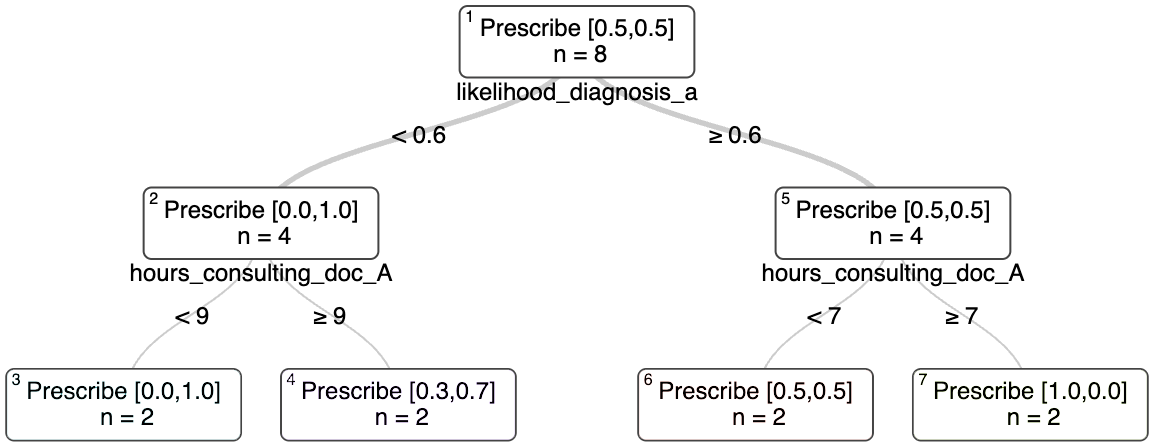}
    \end{minipage}
    
    \vspace{0.5cm}
    
    \textbf{Adaptive Forests} \\[5pt]
    \begin{minipage}{0.48\textwidth}  
        \centering
        \includegraphics[width=1.1\textwidth]{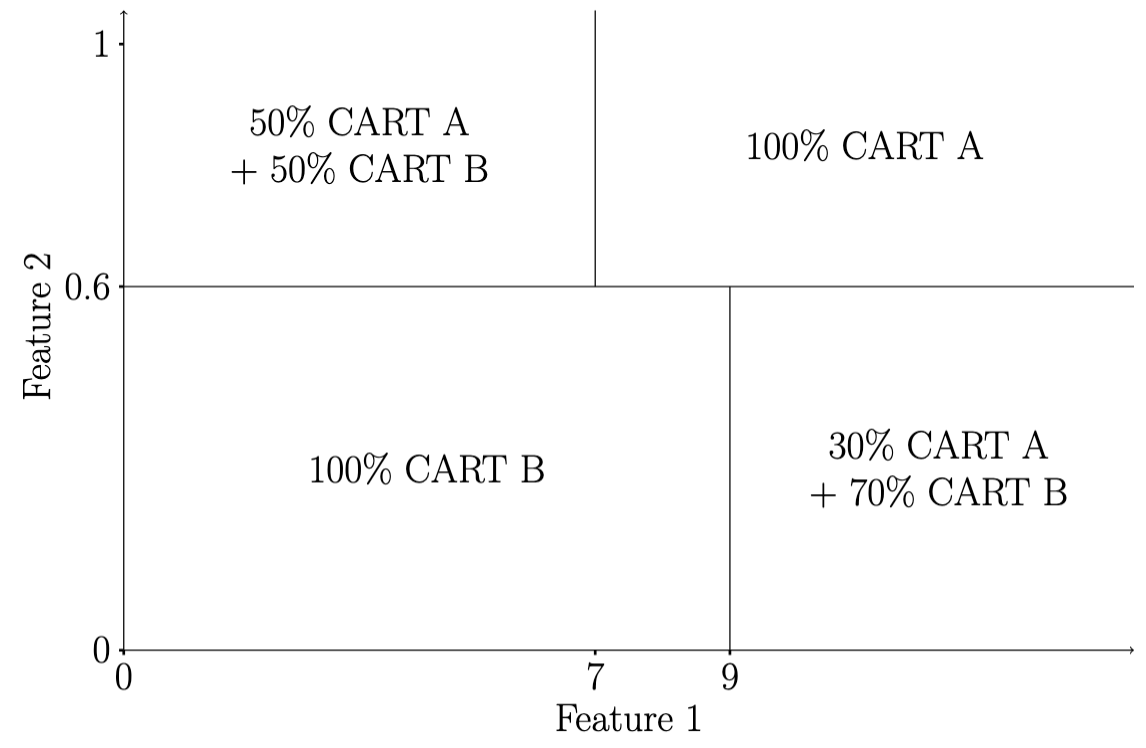}
    \end{minipage}
    \hfill
    \begin{minipage}{0.7\textwidth}  
        \centering
        \includegraphics[width=1\textwidth]{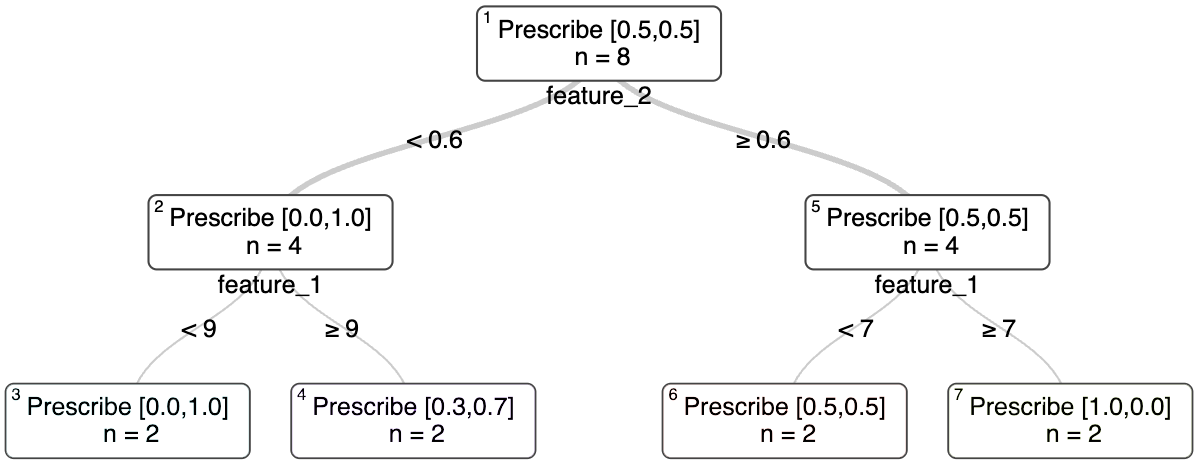}
    \end{minipage}
    \end{adjustwidth} 
    
    \caption{The analogy between human decision making and Adaptive Forests: Consider a binary classification problem where doctors A and B provide opinions on whether a patient should undergo surgery, with doctor A being more specialized in disease a. The decision should weigh their opinions based on two features: (a) time spent consulting doctor A and (b) the probability that this patient has disease a. For example, if a patient has an 80\% probability of disease a and has consulted doctor A for 8 hours, doctor A’s opinion should be prioritized. A policy tree derives this weighting scheme from input features, enhancing interpretability by showing how and when opinions are weighted adaptively. Similarly, for two CART models, A and B, we can learn an adaptive weighting scheme based on feature values to improve predictions. While this example illustrates the core idea behind Adaptive Forests, real-world implementations involve many CART models, a more complex policy tree, and a high-dimensional feature space that cannot be visualized in 2D.}
    \label{fig:group}
    
\end{figure}

In \Cref{fig:group}, we discuss an analogy between weighing the opinion of two doctors and Adaptive Forests.

\subsection{Problem Definition}

Adaptive Forests addresses both binary and multi-class classification problems. For every input data $\xx_i$, each  base CART learner $j\in[m]$ predicts a probability distribution over the  $K$  classes $\ff_j(\xx_i) := (f_{j1}(\xx_i),\,f_{j2}(\xx_i),\,...,\,f_{jK}(\xx_i))$ such that $f_{jk}(\xx_i)\geq0$ and $\sum_{k=1}^K f_{jk}(\xx_i)=1$, where $f_{jk}(\xx_i)$ is the probability that learner $j$ predicts class $k$ on input $\xx_i$. Adaptive Forests further determines instance-specific weights by constructing an Optimal Policy Tree. Instance $\xx_i$ navigates the tree based on splitting rules, landing in exactly one leaf  $l$, which corresponds to a specific weight vector  $\ww_{l(\xx_i)}\in \mathbb{W}$, which is a predefined finite collection of valid weight candidates, i.e. $\mathbb{W} = \{\ww_1,...,\ww_T\}$ for some $T\in \mathbb{Z}_{+}$ s.t. $\ww_t\in \R^m, w_{t,j}\geq 0$, and $\sum_{j=1}^m w_{t,j}=1\,, \forall j\in[m], t\in[T]$. AF applies $\ww_{l(\xx_i)}$  among $m$ base learners and the final prediction is determined by selecting the class  $k$  that maximizes:
\[\AF_k(\xx_i)=\sum_{j=1}^m w_{l(\xx_i),j}f_{jk}(\xx_i).\]

In other words, Adaptive Forests comprises (a) $m$ base CART learners trained on randomly sampled data instances and features, and (b) an Optimal Policy Tree that assigns valid weights from the set $\mathbb{W}$, based on the input \( \mathbf{x}_i \), to aggregate predictions from the \( m \) base learners. Together, Adaptive Forests adaptively weighs predictions from CARTs to enhance out-of-sample performance, evaluated using AUC (Area Under the ROC Curve) for binary classification and One-vs-Rest (OvR) AUC, defined by 
\[\frac{1}{K}\sum_{k=1}^KAUC(\text{class }k, \neg \text{class }k)\footnote{$AUC(\text{class }k, \neg \text{class }k)$ represents the AUC for the binary classification task where class k is considered the positive class, and all other classes are treated as the negative class.},\] for multi-class settings.

\subsection{Related Work}

\paragraph{Optimal Classification Trees}

Optimal Classification Trees (OCT), introduced in \cite{OCT}, is a tree-based machine learning model designed to address the limitations of CART by constructing the entire decision tree in a single step to achieve global optimality. OCT leverages modern Mixed Integer Optimization techniques, which have achieved an 800-billion-fold speedup in recent years, enabling the practical solution of globally optimal decision trees for real-world datasets containing thousands of samples. Synthetic experiments demonstrate that OCTs recover the true decision tree more accurately than heuristic methods, and benchmarks on 53 datasets from the UCI machine learning repository show average absolute improvements in out-of-sample accuracy over CART of $1\text{-}2$\% for binary classification and $3\text{-}5$\% for multi-class classification problems.

\paragraph{Unequal Weighting Variants of Random Forests}
The search for a more effective weighting approach, beyond the standard equal weighting used in Random Forests, has received significant attention. All prior work has employed constant weights, adhering to  $w_{l(\xx_i),j} = w_j$, meaning the weights do not adjust based on  $\xx_i$.

In Trees Weighting Random Forest (TWRF), introduced by \cite{li2010trees}, a weighted voting scheme is proposed for classification tasks, where the weights are determined based on accuracy using the Out-of-Bag (OOB) data. Specifically, let  $OOB_{j}\subset [n]$  represent the subset of data samples not used in training base learner $j$, i.e., the out-of-bag set for tree $j$. In this work, 
\[\overline{w}_j = \frac{1}{|OOB_j|}\sum_{i\in OOB_j} \ind\Big\{\arg\max_{k\in [K]}f_{jk}(\xx_i)=y_i\Big\} ,\]
and $w_j=\frac{\overline{w}_j}{\sum_{r=1}^m \overline{w}_{r}}.$

\cite{kim2011weight} introduced a weighting approach that emphasizes both more accurate classifiers and hard-to-classify instances, i.e., instances that are less frequently classified correctly. The classifier weight vector is computed as follows. First, define a matrix \( M \) of size $n \times m$, where each entry is given by $M_{ij} = \ind \big\{ \arg\max_{k\in [K]} f_{jk}(\xx_i) = y_i \big\}$, indicating whether classifier \( j \) correctly classifies instance \( i \). Next, let \( \mathbf{1}_{ab} \) be an \( a \times b \) matrix consisting entirely of ones, and let \( I_a \) denote the identity matrix of size \( a \times a \). Define the transformed matrix:
\[
\overline{M} := M'(\mathbf{1}_{nm}-M)(\mathbf{1}_{mm}-I_m).
\]
Decompose \( \overline{M} \) into eigenvalues \( \lambda_i \) and eigenvectors \( \uu_i \), for \( i \in [m] \). Let \( r \) be the number of dominant eigenvalues such that \(\lambda_1 = \lambda_2 = \dots = \lambda_r > \lambda_{r+1}\). Finally, compute the classifier weight vector using the dominant eigenvectors:
\[
[w_1, ..., w_m]' = \frac{\left( \sum_{i=1}^{r} \uu_i \uu_i' \right) \mathbf{1}_{m1}}{\mathbf{1}_{1m} \left( \sum_{i=1}^{r} \uu_i \uu_i' \right) \mathbf{1}_{m1}}.
\]

\cite{winham2013weighted} developed Weighted Random Forests (wRF) for binary classification problems, where the weights are determined based on tree-level OOB prediction error $tPE_j$ defined by  
\[tPE_j=\frac{1}{|OOB_j|}\sum_{i\in OOB_j} |f_{j1}(\xx_i)-y_i|.\]
The candidate weight choices are defined as $w_j=\frac{\overline{w}_j}{\sum_{r=1}^m 
\overline{w}_{r}} $ inversely related to $tPE_j$, such as  $\overline{w}_j=1-tPE_j, \exp{\left( \frac{1}{tPE_j}\right)},$ or $(\frac{1}{tPE_j})^\lambda$ with $\lambda=1,2,3,4,5$ etc.

\cite{xuan2018refined} proposed a variant of wRF for binary classification problems by using all training data, including in-bag and out-of-bag data, to compute accuracy measure, which is 
\begin{align*}
    \overline{w}_j &= \sum_{i=1}^n [(f_{jy_i}-(1-f_{jy_i}))\cdot\ind\big\{f_{jy_i} > (1-f_{jy_i})\big\}\\
    &+\alpha\cdot (f_{jy_i}-(1-f_{jy_i}))\cdot\ind \big\{f_{jy_i} < (1-f_{jy_i})\big\}],
\end{align*}
where $\alpha$ is a predefined constant satisfying  $\alpha < 1$  and $w_j=\frac{\overline{w}_j}{\sum_{r=1}^m \overline{w}_{r}}.$

\cite{pham2020cesaro} introduced Cesáro RF (CRF) using the Cesáro average as weights for classification problems. All $m$ base learners are ranked from best to worst following several criteria, for example, the OOB prediction error. For clarity, we redefine $j$ to represent the index of the $j$-th base learner after sorting. Then, $w_j=\frac{\overline{w}_j}{\sum_{r=1}^m 
\overline{w}_{r}}$, where $\overline{w}_j = \sum_{q=j}^m \frac{1}{q}$. 

\cite{zhang2021weighted} proposed a Bayesian-based weighting approach for classification tasks, where the prediction accuracy of base learner \( j \) is determined using the Bayesian formula over a dataset \( S \). The accuracy metric is computed as:  \begin{align*}
    acc_j &= \frac{1}{|S|} \sum_{i\in S} \frac{\Pr[y_i|\argmax_{k\in [K]} f_{jk}(\xx_i)] \Pr[\argmax_{k\in [K]} f_{jk}(\xx_i)]}{\Pr[y_i]}.
\end{align*}
Based on this accuracy, the weight assigned to base learner \( j \) is given by the log-odds transformation: \begin{align*}
    w_j &= \ln \frac{acc_j}{1 - acc_j}.
\end{align*}

Finally, Optimal Weighted Random Forests (OWRF), introduced in \cite{chen2024optimal}, utilizes a Mallows-type weighting criterion to determine optimal weights, primarily for regression tasks. The weights are derived by minimizing a Mallows-type objective function, which carefully balances the trade-off between model fit and complexity:

\begin{align*}
\min_{\mathbf{w}} &\sum_{i=1}^{n} \Big( y_i - \sum_{j=1}^{m} w_j \hat{y}_{ij} \Big)^2 + 2 \sigma^2 \sum_{j=1}^{m} w_j\\
s.t.  & \quad w_j \geq 0 \quad \forall j, \quad \sum_{j=1}^{m} w_j = 1,
\end{align*}
where $\hat{y}_{ij}\in \mathbb{R}$ denotes the prediction made by CART model $j$ for instance $i$, and  $\sigma^2$  represents the estimated noise variance, acting as a regularization term to mitigate model complexity and enhance generalization.

\paragraph{Optimal Policy Trees}

Optimal Policy Trees (OPT), introduced in \cite{OPT}, serves as the engine for the OP2T framework and is therefore central to our Adaptive Forests (AF) approach.

Given observational data $\{\xx_i, t, y_{it}\}_{i=1}^n$, where \( \mathbf{x}_i \) represents the input features, \( t \) is the treatment applied, and \( y_{it} \) is the observed outcome for instance \( i \) under treatment \( t \), OPT performs two key tasks:  
\begin{enumerate}
    \item[(a)] Estimate \( \hat{R}_{it} \), the reward associated with applying treatment \( t \) to input \( \mathbf{x}_i \)\footnote{In certain contexts, such as observational medical data, counterfactual estimation methods may be necessary to compute these rewards. However, AF does not require this, as discussed in Section \ref{subsec: different configurations}.}. 
    \item[(b)] Learn a tree-based policy \( \tau: \mathcal{X} \to \{t_1, ..., t_T\} \) that partitions the feature space into meaningful groups of similar instances and assigns the optimal treatment to each partition, thereby maximizing the overall reward:
    \[\max_{\tau(\cdot)}\sum_{i=1}^n \sum_{t=t_1}^{t_T}\ind\big \{\tau(\xx_i)=t\big\}\hat{R}_{it}+\lambda\cdot numsplits(\tau),\]
where the number of splits in $\tau$ is regulated by the hyperparameter $\lambda\leq 0$ to avoid over-fitting, along with the constraints that $\tau$ belongs to the class of decision trees with a predefined maximum depth and that each leaf contains at least $c_{min}$ samples.
\end{enumerate}

The OPT formulation enables the construction of interpretable, tree-based policies that effectively capture nonlinear interactions and segment the feature space. Unlike traditional greedy heuristics, OPT employs a global optimization approach, leveraging coordinate descent to jointly optimize the tree structure and treatment assignments. Built upon the Optimal Trees framework from \cite{bertsimas2019machine}, this method significantly improves both performance and interpretability compared to heuristic-based policy trees.

\paragraph{Optimal Predictive-Policy Trees (OP2T)}\label{OP2T}

The Optimal Predictive-Policy Trees (OP2T) framework, introduced by \cite{OP2T}, reframes prediction problems as prescription problems by optimizing how to ensemble multiple base learners. 

Given a prediction task with data $\{\xx_i, y_i\}_{i=1}^n$, we can train $m$  base learners and ensemble them for improved performance. The challenge is to identify the optimal approach for combining these predictions, modeled as assigning a weight vector $\ww$, analogous to making treatment decisions in a prescription problem framework. The treatments \( t, t = 1, \dots, T \), correspond to weight vectors \( \mathbb{W} = \{ \mathbf{w}_1, \dots, \mathbf{w}_T \} \), where each weight vector $\mathbf{w}_t = (w_{t1}, \dots, w_{tm})$ satisfies $w_{tr} \geq 0$ and $\sum_{r=1}^{m} w_{tr} = 1$. Here, \( w_{tr} \) represents the weight assigned to model \( r \) under treatment \( t \).

In addition to treatments, rewards $R_{i,t}$ for data $\xx_i$ under treatment $t$ can be computed using the ground truth $y_i$ and ensembled prediction $\sum_{j=1}^m w_{tj}\ff_j(\xx_i)$. An example of such a reward for binary classification is

\[R_{i,t} = \frac{1}{|y_i-\sum_{j=1}^m w_{tj}f_{j1}(\xx_i)|+1},\]
which suggests that a lower error corresponds to a higher reward. It is important to note that no counterfactual estimation is required here, as the outcome of applying treatment  $t$  can be directly computed.

Building upon OPT, OP2T segments the feature space based on model effectiveness and identifies subpopulations where specific models excel or underperform, enabling more informed deployment decisions. Building upon globally optimized prescriptive trees, OP2T surpasses traditional heuristic-based approaches by offering greater interpretability and adaptability across diverse datasets. Empirical studies on real-world regression and classification tasks demonstrate its strong performance, providing valuable insights into optimal model usage.

\subsection{Contributions}

The main contributions of our work are as follows:
\begin{itemize}
    \item We present Adaptive Forests, a novel tree-based model for general classification tasks that delivers superior out-of-sample performance compared to Random Forests, XGBoost, and other weighted random forest algorithms, such as wRF. In addition to achieving better evaluation metrics, AF stands out for its robust performance 
    (AF is  one of the two top-performing models in AUC for binary classification  in 10 out of 14 instances and in OvR AUC for  multi-class classification in 5 out of 7 instances)  and reduced model complexity
    (AF uses a maximum of 50 and 100 CART trees for binary and multi-class classification, respectively, compared to 1000 CART trees used by other methods). These underscore the benefits of adaptively assigning unequal weights to base learners based on input features, in contrast to the equal weighting or majority voting strategies used by Random Forests.

    \item Adaptive Forests achieves effective adaptiveness through two key mechanisms: (a) leveraging the OP2T framework to derive a structured policy for assigning unequal weights, and (b) integrating Mixed Integer Optimization algorithms to dynamically generate and select new weights, refining the weighting policy through iterative updates. The second mechanism significantly enhances Adaptive Forests’ performance, extending beyond the OP2T framework, which is limited to a finite set of pre-defined weights for ensemble weighting.
    
    \item We introduce various configurations for OPT training, offering flexibility and greater potential for improving AF's performance.
\end{itemize}

\paragraph{Organization of the paper.} The rest of the paper is organized as follows. In Section \ref{sec: main AF algo}, we introduce the proposed Adaptive Forests algorithm, describing its overall framework, outlining the optimization methods employed to adaptively generate new weight candidates for improved performance, and examining various configuration and hyperparameter options. In Section \ref{sec:experiments}, we present the experimental results of applying the Adaptive Forests algorithm to multiple real-world datasets, addressing both binary and multi-class classification tasks. We compare the performance of Adaptive Forests against Random Forests, XGBoost, and Weighted Random Forests (wRF). Using metrics such as (OvR) AUC and the number of wins, we demonstrate that AF achieves superior and robust performance while requiring fewer underlying CARTs. Finally, Section \ref{sec:conclusion} concludes the paper with a summary of the key contributions.

\section{Adaptive Forests}\label{sec: main AF algo}

In this section, we delve into the proposed Adaptive Forests (AF) algorithm, beginning with a high-level outline of its overall framework.

Adaptive Forests is composed of (a) $m$  independent CART base learners, each trained on randomly sampled data instances and feature subsets, and (b) an OPT that assigns a customized weight vector to each input $\xx$. The overall training pipeline follows these steps:
\begin{enumerate}
    \item \textbf{Dataset Partitioning:} Partition the dataset according to \Cref{table:data division}. We reserve 20\% of the data as \texttt{data\_test} for final performance assessment. From the remaining 80\%, 15\% is allocated to \texttt{data\_val} for selecting the best configuration, while the rest is divided in a 60:40 ratio between \texttt{data\_single} (used for CART training) and \texttt{data\_opt} (used for OPT training).

    \item \textbf{Base Learner Training:} Train $m$ CART base learners using the \texttt{data\_single} subset.
    \item \textbf{Weight Initialization:} Initialize the set of weight candidates $\mathbb{W}$ based on the methods detailed in \Cref{appendix A: weight init}.
    \item \textbf{Configuration Selection:} Select the optimal configuration using \texttt{data\_val} in the first iteration, detailed in \Cref{subsec: different configurations}. 
    \item \textbf{Iterative Optimization:} Iteratively generate new weight candidates to augment $\mathbb{W}$ using Mixed-Integer Optimization (MIO) and train a new OPT with the updated set, repeating the process until convergence or a maximum of 10 iterations, detailed in \Cref{subsec: MIO}. 
    \item \textbf{Final Evaluation:} Evaluate and report performance on the \texttt{data\_test} at the end.
\end{enumerate}

\begin{table}[h!]
\centering
\begin{tabular}{|l|c|l|}
\hline
\textbf{Data Subset} & \textbf{Portion} & \textbf{Purpose} \\ \hline
\texttt{data\_single} & \( 0.8 \times 0.85 \times 0.6 \) & Used to train CARTs \\ \hline
\texttt{data\_opt} & \( 0.8 \times 0.85 \times 0.4 \) & Used to train OPT \\ \hline
\texttt{data\_val} & \( 0.8 \times 0.15 \) & Used to select the best configuration for OPT training \\ \hline
\texttt{data\_test} & \( 0.2 \) & Used to evaluate performance only at the very end \\ \hline
\end{tabular}
\caption{Data Subsets and Their Uses.}
\label{table:data division}
\end{table}

Beyond the core structure of this algorithm, two additional components play a crucial role in achieving the superior performance of Adaptive Forests (AF) and will be examined in this section. The first key component focuses on selecting an effective set of weight candidates from the full probability simplex and refining them throughout training using MIO-based weight generation techniques. Unlike the OP2T framework, which restricts weight selection to a predefined finite set, our approach leverages MIOs to dynamically explore and adaptively update weight candidates. This flexibility allows for a more customized model, significantly enhancing the overall performance of Adaptive Forests. The second critical aspect involves optimizing the algorithm’s potential by carefully selecting configuration choices, such as defining the reward matrix and identifying the most informative subset of training data for OPT training. Given the vast range of possible configurations, it is essential to determine the optimal setup for each dataset to maximize performance. Together, the adaptability of Adaptive Forests is driven by three fundamental factors: (a) unlike traditional methods that rely on fixed weights, Adaptive Forests assigns weights $\ww_{l(\xx_i)}$  that dynamically adjust based on the input  $\xx_i$; (b) the set of weight candidates  $\mathbb{W}$  is iteratively updated throughout training to better align with the specific problem; and (c) during the first training iteration, an optimal configuration for OPT training is selected to best fit the dataset characteristics.

\subsection{ Optimization Ideas}\label{subsec: MIO}

The core of Adaptive Forests lies in employing optimization techniques to iteratively and adaptively refine the set of candidate weights \( \mathbb{W} \) from which OPT selects and assigns weights. This process involves (a) generating high-quality weight candidates and (b) identifying the most effective ones to update  \( \mathbb{W} \). While the OP2T framework serves as a foundation, AF goes beyond it by dynamically optimizing weight selections to enhance overall performance.

Given the current set of candidate weights \( \mathbb{W} = \{\ww_1, ..., \ww_T\} \), we outline how to generate the new weight \( \ww_{T+1} \) based on MIO. The objective is to identify a weight vector that minimizes the overall error. Taking  $\Ic$  as either the complete set of instances in \texttt{data\_opt} or the subset corresponding to each leaf of the current OPT. For binary classification, 
\begin{subequations} \label{eq:chunk1}
\begin{align}
\ww_{T+1} = \argmin_{\mathbf{w}} \hspace{0.2cm} &\sum_{i\in\Ic} \Big( \ind\Big\{\sum_{j=1}^m w_{j}f_{j1}(\xx_i)\geq 0.5\Big\}-y_i\Big)^2\\
s.t. \hspace{0.2cm}& \mathbf{w}\in \R^m_{\geq 0},\, \sum_{j=1}^m w_j=1, 
\end{align}
\end{subequations} 
and for multi-class classification, 
\begin{subequations} \label{eq:chunk3}
\begin{align}
\ww_{T+1} = \argmax_{\mathbf{w}} \hspace{0.2cm} &\sum_{i\in\Ic}
\ind \Big\{\argmax_{k\in[K]}\sum_{j=1}^m w_{j}f_{jk}(\xx_i) = y_i\Big\}
\\
s.t. \hspace{0.2cm}& \mathbf{w}\in  \R^m_{\geq 0}, \,\sum_{j=1}^m w_j=1.
\end{align}
\end{subequations} 
To solve problems \eqref{eq:chunk1} and \eqref{eq:chunk3} using MIO, they must first be reformulated with standard MIO techniques. A detailed transformation into a programmable format for an optimization solver is available in \Cref{appendix B: mio linear}.

Building on the vanilla optimization problems \eqref{eq:chunk1} and \eqref{eq:chunk3}, we introduce additional constraints to guide the search for optimal weight candidates, balancing the need for both exploration and exploitation:

\begin{itemize}
    \item \textbf{Exploration:} Encourage the generation of a new weight vector \( \ww \) that is sufficiently different from all existing candidates:
    \[
    \|\mathbf{w}-\mathbf{w}'\|_2 \geq \textit{min\_gap}, \quad \forall \mathbf{w}' \in \mathbb{W},
    \]
    where \( \textit{min\_gap} \) is a positive hyperparameter. Note that these constraints are nonconvex quadratic constraints, and appropriate attributes should be configured for the MIO solver accordingly.

    \item \textbf{Exploitation:} Ensure that the new weight vector \( \ww \) remains close to at least one historically used weight:
    \begin{align*}
        &s_{\ww '} \in \{0,1\}, \quad \forall \mathbf{w}' \in \textit{hist\_}\mathbb{W},\\
        &\sum_{\mathbf{w}'\in \textit{hist\_}\mathbb{W}} s_{\ww'} \leq |\textit{hist\_}\mathbb{W}| - 1,\\
        &\|\ww' - \ww\|_2 \leq 10 \cdot s_{\ww '} + \textit{max\_gap}, \quad \forall \mathbf{w}' \in \textit{hist\_}\mathbb{W},
    \end{align*}
    where \( \textit{max\_gap} \) is a positive hyperparameter, and \( \textit{hist\_}\mathbb{W} \) denotes the collection of all weight vectors that have been selected by at least one OPT in previous iterations. Instead of considering the full \(\mathbb{W} \), we restrict our search to  \( \textit{hist\_}\mathbb{W} \)  as part of an exploitation strategy, which focuses on refining previously selected weight vectors to uncover further improvements. This approach assumes that if a weight vector has never been chosen before, exploring its surroundings is unlikely to yield meaningful gains. Thus, exploitation is guided by past selections, ensuring a more targeted and potentially efficient optimization process. In this formulation, the binary variable \( s_{\mathbf{w}'} \) indicates whether slack is allowed for a given weight vector \( \mathbf{w}' \in \textit{hist\_}\mathbb{W} \)—that is, whether the generated vector \( \mathbf{w} \) is permitted to deviate from \( \mathbf{w}' \). The second constraint ensures that slack can be applied to at most all but one of the \( \mathbf{w}' \) vectors, requiring that at least one remain close to the generated solution. The third constraint enforces this condition: if \( s_{\mathbf{w}'} = 1 \), the constraint is effectively relaxed; if \( s_{\mathbf{w}'} = 0 \), then the Euclidean distance between \( \mathbf{w} \) and \( \mathbf{w}' \) must be less than or equal to the specified threshold \( \textit{max\_gap} \).
    
\end{itemize}

To mitigate overfitting and reduce computational costs, it is essential to regulate the size of \( \mathbb{W} \). Instead of incorporating all newly generated weights without selection, the following MIO formulation allows for choosing the most promising \( k \) weights from the set of newly generated weight candidates \( new\_\mathbb{W} \). Let \( \Lc \) represent the set of all leaves in the current OPT:

\begin{align}\label{refer error}
    \max_{\mathbf{m}, \mathbf{s}} \quad & \sum_{l \in \Lc}\sum_{\ww\in new\_\mathbb{W}} m_{l \ww} \sum_{q\in l} \ind  \Big\{ \argmax_{k\in[K]}\sum_{j=1}^m w_{j}f_{jk}(\xx_q)  = y_q \Big\} \hypertarget{eq: select k}\\ \nonumber
    \text{s.t.} \quad & \mathbf{m} \in \{0,1\}^{|\Lc|\times |new\_\mathbb{W}|},\\ \nonumber
    & \mathbf{s} \in \{0,1\}^{|new\_\mathbb{W}|},\\ \nonumber
    & \sum_{\ww} m_{l \ww} = 1, \quad \forall l\in \Lc, \\\nonumber
    & \sum_{\ww\in new\_\mathbb{W}} s_{\ww} \leq k, \\\nonumber
    & m_{l \ww} \leq s_{\ww}, \quad \forall l\in\Lc, \: \ww\in new\_\mathbb{W}.
\end{align}

\subsection{Configuration and Hyperparameter Choices}\label{subsec: different configurations}

The Adaptive Forests algorithm can make some choices that affect its overall performance. In this section, we discuss these choices, among which the most suitable configuration for each dataset is determined using validation data \texttt{data\_val} during the initial iteration.

\paragraph{Choice of Reward Matrix.}The reward matrix \( R \in \mathbb{R}^{n \times |\mathbb{W}|} \) quantifies the effectiveness of each weight vector \( \ww \in \mathbb{W} \) in relation to a given input \( \xx_i \) for $i\in[n]$. It serves as a critical component in guiding the construction of OPT, ensuring that expected rewards are maximized for both observed and unobserved instances. Specifically, in the Adaptive Forests framework, \( R_{i\ww} \) quantifies the likelihood that applying \( \ww \) to weigh predictions from the \( m \) base learners will result in a correct prediction for \( \xx_i \). Below, we outline the various choices we used to compute the reward matrix.

Let \( P(\xx_i, \ww) \) represent the predicted probability distribution across \( K \) classes for the input \( \xx_i \), obtained by aggregating predictions from \( m \) base CART learners using the weight vector \( \ww \), i.e. 
\[P(\xx_i, \ww) := \begin{pmatrix}
\sum_{j=1}^m w_{j}f_{j1}(\xx_i)\\
\sum_{j=1}^m w_{j}f_{j2}(\xx_i)\\
...\\
\sum_{j=1}^m w_{j}f_{jK}(\xx_i)\\
\end{pmatrix}.\]

\begin{itemize}
    \item Hard reward, i.e., $R_{MIS}$ from \cite{OP2T}: 
    \[R^{hard}_{i\ww} = \ind \Big\{y_i = \operatorname*{argmax}_{k\in [K]} \:\sum_{j=1}^m w_{j}f_{jk}(\xx_i)\Big\}.\]
    \item Soft reward:
    \[R^{soft}_{i\ww} = \sum_{j=1}^m w_{j}f_{jy_i}(\xx_i).\]
    \item Threshold soft reward parameterized by $\alpha\in\R_{+}:$
    \[R^{threshold\_soft}_{i\ww} = R^{soft}_{i\ww} * \ind \Big\{R^{soft}_{i\ww} \geq \alpha\Big\}.\]

    \item Euclidean distance reward:
    
    \[R^{ED}_{i\ww} = \frac{1}{ \|P(\xx_i,\ww)-\mathbf{e}_{y_i}\|+1} .\]
    
    \item Kullback–Leibler (KL) divergence reward:

    \[R^{KL}_{i\ww} = \frac{1}{D_{\text{KL}}( P(\xx_i,\ww)\| \mathbf{e}_{y_i})+1}.\]

    \item Negative cross-entropy reward:

    \[R^{NCE}_{i\ww} = -H(\mathbf{e}_{y_i}, P(\xx_i,\ww)). \footnote{ $H(p, q) := -\sum_{i} p(i) \log q(i)$ for discrete probability distribution.}\]

\end{itemize}

It is important to note that computing the entire reward matrix does not require any counterfactual estimation, given that we have access to \( \mathbb{W} \), \( y_i \), and \( f_{jk}(\xx_i) \) for \( j \in [m], k \in [K] \) and each instance \( i \) from \texttt{data\_opt}.

\paragraph{Choice of  Input Features for OPT.}

The input feature selection for OPT defines the basis on which it assigns weights to maximize expected rewards. In addition to using the original input \( \xx_i \), we can extend it by including \( f_{jk}(\xx_i) \) for \( j \in [m], k \in [K] \), representing predictions from all base learners.

Referring to our earlier example, the decision of which doctor’s opinion to prioritize can be influenced by the patient’s characteristics \( \xx_i \), such as the patient's history of frequent visits to a specific doctor. Another consideration is the confidence level of each doctor’s prediction, which can be inferred from \( f_{j1}(\xx_i) \), where values close to \( 0 \) or \( 1 \) indicate high confidence, while those near \( 0.5 \) suggest uncertainty.

\paragraph{Choice of Which Data from \texttt{data\_opt} to Use.}

We can refine the training dataset for OPT by selecting a subset of \texttt{data\_opt}, excluding instances that are either uninformative or misleading, irrespective of the assigned weights. This method is particularly relevant for binary classification tasks. Consider the following scenario: for a data instance \( \xx_i \) with true label \( y_i = 1 \), if \( f_{j1}(\xx_i) > 0.5, \:\forall j \in [m] \) or \( f_{j1}(\xx_i) < 0.5, \:\forall j \in [m] \), then the weighted result satisfies  
\[
\sum_{j=1}^m w_j f_{j1}(\xx_i) > 0.5 \quad \text{or} \quad \sum_{j=1}^m w_j f_{j1}(\xx_i) < 0.5
\]  
for any \( \mathbf{w} \in \mathbb{R}^m_{\geq 0} \) such that \( \sum_{j=1}^m w_j = 1 \). In the first case, the prediction is deterministically correct, while in the second, it is deterministically incorrect, both independent of \( \ww \). As a result, retaining such instances may not contribute to OPT’s ability to learn an effective weight assignment. Worse, instances from the latter scenario could introduce misleading noise into the training process. Building on this concept, we can select the actual input data for OPT training from the following options:  
\begin{enumerate}
    \item[(a)] All of \texttt{data\_opt}.  
    \item[(b)] Only the non-deterministic portion of \texttt{data\_opt}.  
    \item[(c)] With the deterministic and correct portion removed.  
    \item[(d)] With the deterministic and incorrect portion removed.  
\end{enumerate}

\paragraph{Remark:} Experiments on real-world datasets reveal that no single configuration consistently outperforms others across all datasets. Which combination performs better often depends on the specific task at hand. Leveraging \texttt{data\_val} to identify the optimal configuration is generally the most effective strategy.

\section{Performance in Real World Data}\label{sec:experiments}

The Adaptive Forests algorithm is evaluated using real-world datasets from the UCI Machine Learning Repository (\cite{UCIdata}) and the Boston Medical Center (BMC) Diabetes Dataset (\cite{bmc_diabetes_dataset}). We evaluate its performance against Random Forests, XGBoost, and a weighted random forest algorithm referred to as wRF. Both Random Forests and XGBoost are implemented using Interpretable AI's package (\url{https://docs.interpretable.ai/stable/IAI-Python/reference/#XGBoostLearner}), with hyperparameters optimized through grid search on validation datasets, considering up to 1000 trees. The wRF algorithm utilizes CARTs trained with Interpretable AI’s package, while the custom unequal weighting mechanism represents our own implementation. It is worth noting that the original wRF algorithm is designed for binary classification only, and we extend it to support multi-class classification to compare with AF. All experiments are conducted using CART with depths of 10 and 100 to examine the impact of the individual model’s depth on performance. The number of underlying CARTs of AF is capped at 50 for binary classification and 100 for multi-class classification, providing sufficient capacity for strong performance while minimizing computation time. The same experiments have been conducted, substituting CART with OCT, but no significant improvements were observed. To ensure a clearer comparison between AF and RF, XGBoost, and wRF, all of which use the same underlying models, we report results exclusively with CART as the base learner.

In Tables \ref{tab:bin_10}, \ref{tab:bin_100}, \ref{tab:multi_10}, and \ref{tab:multi_100}, we report the computational performance of RF, XGBoost, wRF, and AF for both binary and multi-class classification tasks, evaluating base learners with depths of 10 and 100.  Here,  $n$  represents the total number of data instances in the dataset,  $p$  denotes the number of features, and  $K$  indicates the number of classes in multi-class classification problems.

\newpage
\begin{table}[h!]
\centering
\begin{tabular}{lcccccc}
\toprule
\textbf{Dataset Name} & \textbf{n} & \textbf{p} & \textbf{RF} & \textbf{XGBoost} & \textbf{wRF} & \textbf{AF} \\
\midrule
blood-transfusion & 748 & 5 & 0.6980 & 0.6770 & 0.7417 & 0.7535 \\
breast-cancer & 286 & 10 & 0.6775 & 0.6193 & 0.7059 & 0.7559 \\
breast-cancer-prognostic & 198 & 35 & 0.6111 & 0.7311 & 0.6222 & 0.7556 \\
diabetes-bin-500 & 500 & 31 & 0.8474 & 0.8624 & 0.8368 & 0.8299 \\
echocardiogram & 74 & 12 & 0.9900 & 0.9940 & 1.0000 & 1.0000 \\
haberman-survival & 306 & 4 & 0.6542 & 0.6931 & 0.7188 & 0.6840 \\
house-votes & 435 & 17 & 0.9928 & 0.9900 & 0.9889 & 0.9917 \\
indian-liver-patient & 583 & 11 & 0.6870 & 0.7711 & 0.7733 & 0.7585 \\
monks-problems-2 & 601 & 7 & 0.9949 & 0.9947 & 0.8873 & 0.9559 \\
planning-relax & 182 & 13 & 0.4885 & 0.5315 & 0.3962 & 0.5654 \\
spect-heart & 267 & 23 & 0.7868 & 0.7911 & 0.8690 & 0.8571 \\
spectf-heart & 267 & 45 & 0.8485 & 0.8442 & 0.8398 & 0.8463 \\
statlog-german-credit & 1000 & 25 & 0.7777 & 0.8090 & 0.7957 & 0.8096 \\
wdbc & 569 & 32 & 0.9894 & 0.9993 & 0.9943 & 0.9943 \\
\midrule
\textbf{Average} &  &  & 0.7888 & 0.8077 & 0.7978 & 0.8256 \\
\textbf{Number of Wins} &  &  & 3 & 2 & 4 & 6 \\
\bottomrule
\end{tabular}
\caption{AUC for Binary Classification Tasks Using CART with Depth 10.}
\label{tab:bin_10}
\end{table}

\begin{table}[h!]
\centering
\begin{tabular}{lcccccc}
\toprule
\textbf{Dataset Name} & \textbf{n} & \textbf{p} & \textbf{RF} & \textbf{XGBoost} & \textbf{wRF} & \textbf{AF} \\
\midrule
blood-transfusion & 748 & 5 & 0.6896 & 0.6613 & 0.7390 & 0.7538 \\
breast-cancer & 286 & 10 & 0.6529 & 0.6219 & 0.7059 & 0.7074 \\
breast-cancer-prognostic  & 198 & 35 & 0.5944 & 0.7311 & 0.6222 & 0.7000 \\
diabetes-bin-500 & 500 & 31 & 0.8406 & 0.8568 & 0.8368 & 0.8299 \\
echocardiogram & 74 & 12 & 0.9900 & 0.9940 & 1.0000 & 1.0000 \\
haberman-survival & 306 & 4 & 0.6556 & 0.7083 & 0.7188 & 0.6840 \\
house-votes & 435 & 17 & 0.9922 & 0.9900 & 0.9218 & 0.9917 \\
indian-liver-patient & 583 & 11 & 0.7187 & 0.8025 & 0.7736 & 0.7585 \\
monks-problems-2 & 601 & 7 & 0.9948 & 0.9947 & 0.8873 & 0.9537 \\
planning-relax & 182 & 13 & 0.4819 & 0.5392 & 0.3885 & 0.5654 \\
spect-heart & 267 & 23 & 0.7716 & 0.7781 & 0.8690 & 0.8571 \\
spectf-heart & 267 & 45 & 0.8452 & 0.8442 & 0.8398 & 0.8463 \\
statlog-german-credit & 1000 & 25 & 0.7562 & 0.8007 & 0.7945 & 0.8086 \\
wdbc & 569 & 32 & 0.9894 & 0.9993 & 0.9889 & 0.9943 \\
\midrule
\textbf{Average} & & & 0.7838 & 0.8087 & 0.7919 & 0.8179 \\
\textbf{Number of Wins} & & & 2 & 4 & 3 & 6 \\
\bottomrule
\end{tabular}
\caption{AUC for Binary Classification Tasks Using CART with Depth 100.}
\label{tab:bin_100}
\end{table}

\begin{table}[h!]
\centering
\begin{tabular}{lccccccc}
\toprule
\textbf{Dataset Name} & \textbf{n} & \textbf{p} & \textbf{K} & \textbf{RF} & \textbf{XGBoost} & \textbf{wRF} & \textbf{AF} \\
\midrule
balance-scale & 625 & 5 & 3 & 0.8465 & 0.9499 & 0.8794 & 0.9010 \\
contraceptive-method-choice & 1473 & 10 & 3 & 0.7267 & 0.6836 & 0.7417 & 0.7369 \\
diabetes-multi-500 & 500 & 31 & 12 & 0.6882 & 0.8316 & 0.9032 & 0.8991 \\
diabetes-multi-5000 & 5000 & 31 & 12 & 0.8932 & 0.9285 & 0.9376 & 0.9383 \\
heart-disease-cleveland & 303 & 14 & 5 & 0.7161 & 0.7542 & 0.7351 & 0.7845 \\
teaching-assistant & 151 & 6 & 3 & 0.7633 & 0.8267 & 0.8150 & 0.8067 \\
yeast & 1484 & 10 & 10 & 0.9004 & 0.8828 & 0.8890 & 0.8812 \\
\midrule
\textbf{Average} &  &  &  & 0.7906 & 0.8367 & 0.8430 & 0.8497 \\
\textbf{Number of Wins} &  &  &  & 1 & 2 & 2 & 2 \\
\bottomrule
\end{tabular}
\caption{One-vs-Rest AUC for Multi-class Classification Tasks Using CART with Depth 10.}
\label{tab:multi_10}
\end{table}

\begin{table}[h!]
\centering
\begin{tabular}{lccccccc}
\toprule
\textbf{Dataset Name} & \textbf{n} & \textbf{p} & \textbf{K} & \textbf{RF} & \textbf{XGBoost} & \textbf{wRF} & \textbf{AF} \\
\midrule
balance-scale & 625 & 5 & 3 & 0.8464 & 0.9538 & 0.8795 & 0.9011 \\
contraceptive-method-choice & 1473 & 10 & 3 & 0.6684 & 0.6756 & 0.7407 & 0.7305 \\
diabetes-multi-500 & 500 & 31 & 12 & 0.7124 & 0.8369 & 0.9011 & 0.8996 \\
diabetes-multi-5000 & 5000 & 31 & 12 & 0.8587 & 0.9216 & 0.9368 & 0.9384 \\
heart-disease-cleveland & 303 & 14 & 5 & 0.7102 & 0.7620 & 0.7386 & 0.7810 \\
teaching-assistant & 151 & 6 & 3 & 0.7500 & 0.8283 & 0.8000 & 0.8150 \\
yeast & 1484 & 10 & 10 & 0.8973 & 0.8763 & 0.8808 & 0.8765 \\
\midrule
\textbf{Average} &  &  &  & 0.7776 & 0.8363 & 0.8396 & 0.8489 \\
\textbf{Number of Wins} &  &  &  & 1 & 2 & 2 & 2 \\
\bottomrule
\end{tabular}
\caption{One-vs-Rest AUC for Multi-class Classification Tasks Using CART with Depth 100.}
\label{tab:multi_100}
\end{table}

\newpage

From these tables, we can make several key observations:
\begin{itemize}
    \item {\bf Performance and its Robustness for Binary Classification.} Adaptive Forests (AF) consistently outperforms RF, XGBoost, and wRF in terms of both average AUC and the number of winning instances for binary classification. This trend remains consistent across CART depths of 10 and 100.   Regarding robustness in their performance, AF is one of the two top-performing models in AUC for binary classification in 10 out of 14 instances.
    
    \item  {\bf Performance and its Robustness for Multi-class Classification.} AF achieves the highest average One-vs-Rest (OvR) AUC, although the number of wins is comparable across the four algorithms, for multi-class classification. This pattern remains consistent between CART depths of 10 and 100, with AF’s advantage in average OvR AUC becoming more pronounced at depth 100. 
    Regarding robustness in their performance,  AF is one of the two top-performing models in OvR AUC   in 5 out of 7 instances.

    \item {\bf Reduced Model Complexity.}  It is worth emphasizing that AF uses up to 50 underlying CARTs for binary classification and up to 100 for multi-class classification to achieve the reported performance, in contrast to the benchmark algorithms, which utilize up to 1000 base learners. Despite employing significantly fewer base learners, AF consistently delivers superior performance in terms of average (OvR) AUC and the number of wins. While multiple OPTs are trained iteratively and adaptively, only the best one is retained, which tends to be relatively shallow. Consequently, the overall model complexity of AF is significantly lower than that of the benchmark algorithms without any compromise—in fact, it offers enhanced performance.

    \item In summary, Adaptive Forests (AF) consistently outperforms Random Forests (RF), XGBoost, and weighted Random Forests (wRF) in both binary and multi-class classification tasks. AF excels not only in achieving superior evaluation metrics but also in their performance robustness and reduced model complexity. This highlights the advantage of adaptively assigning unequal weights to base learners based on input features, as opposed to the equal weighting or majority voting approach employed in RF.

\end{itemize}

\section{Conclusions}\label{sec:conclusion}

This paper sets out to introduce a novel tree-based model, Adaptive Forests, designed for general classification tasks. This model enhances Random Forests by incorporating the ability to assign unequal weights to ensemble base learners’ predictions adaptively for each input data point. The key techniques employed include the interpretable and adaptive ensemble policy learning framework, OP2T, along with MIOs to iteratively generate and update ensemble weights. Tested on real-world datasets, Adaptive Forests demonstrates superior performance compared to Random Forests, XGBoost, and other weighted random forest algorithms. Given its generalizability and enhanced predictive power, Adaptive Forests has the potential to become a new go-to model within the tree-based family, with applications across various industries.

\acks{The authors acknowledge the MIT SuperCloud and Lincoln Laboratory Supercomputing
Center for providing high-performance computing resources that have contributed to the
research results reported in this paper.}

\vskip 0.2in
\bibliography{main}

\newpage
\appendix
\section{Initialization of Weight Candidate Set}\label{appendix A: weight init}

We initialize the set of ensemble weight candidates $\mathbb{W}$ by uniformly sampling a specified number $|\mathbb{W}|$ of points from the probability simplex $\{ \mathbf{p} \in \mathbb{R}^m \mid p_i \geq 0, \sum_{i=1}^{m} p_i = 1 \}$. 

Alternatively, we can initialize  $\mathbb{W}$  with a warm start as follows:
\begin{enumerate}
    \item Uniform weighting $\ww^1$: $\ww^1_i=\frac{1}{m} ,\: \forall i\in[m]$.
    \item Single model selected $\ww^j$: $\ww^j_j=1$ and $\ww^j_i=0 ,\: \forall i\neq j, \: j\in [m]$.
    \item Two models selected $\ww^{jk}: \ww^{jk}_j=\ww^{jk}_k=\frac{1}{2}$ and $\ww^{jk}_i=0,\:\forall i\neq j,k, \:$ for all pairs  (j, k)  chosen from $m$  elements. 
    \item Up to $q$ models are selected, each receiving uniform weight, for every $q$-element subset selected from  $m$ elements. 
\end{enumerate}

The first method allows explicit control over the size of $\mathbb{W}$ and introduces greater randomness, which can help mitigate the risk of getting trapped in local optima during subsequent weight generation MIO’s optimization process. In contrast, the second method offers the advantage of producing more interpretable and readable weights; however, its drawback is that the size of $\mathbb{W}$ grows combinatorially with the number of base learners $m$.

\section{Transforming into Programmable MIO Formulations}\label{appendix B: mio linear}

To make Eq.\eqref{eq:chunk1} programmable, it can be transformed into the following:
\begin{subequations} \label{eqn: vanilla binary}
\begin{align}
\min_{\mathbf{w}} \hspace{0.2cm} &\sum_{i\in\Ic}(k_i-y_i)^2\\
s.t. \hspace{0.2cm}& M\cdot (1-k_i)+\sum_{j=1}^m w_{j}f_{j1}(\xx_i)\geq 0.5,\:\forall i\in \mathcal{I},\\
&\sum_{j=1}^m w_{j}f_{j1}(\xx_i)+\epsilon\leq 0.5+M\cdot k_i,\:\forall i\in \mathcal{I},\\
&\mathbf{w}\in \R^m_{\geq 0},\\ &\sum_{j=1}^m w_j=1,\\
&\mathbf{k}\in\{0,1\}^{|\Ic|},
\end{align}
\end{subequations}
where $M=2$ is big enough and $\epsilon>0$ to be small.

We transform Eq.\eqref{eq:chunk3} into an equivalent formulation suitable for implementation using the following techniques:

We define \[b(i,q):=\ind \Big\{\sum_{j=1}^m w_{j}f_{jy_i}(\xx_i)\geq \sum_{j=1}^m w_{j}f_{jq}(\xx_i)\Big\}, \:\forall i\in \mathcal{I}, q\in [K],\]
which can be represented using the following constraints in the MIO formulation:

\begin{align*}
    &b(i,q)\in\{0,1\},\\ 
    &M\cdot (1-b(i,q))+\sum_{j=1}^m w_{j}f_{jy_i}(\xx_i) \geq \sum_{j=1}^m w_{j}f_{jq}(\xx_i),\\
    &\sum_{j=1}^m w_{j}f_{jy_i}(\xx_i)+\epsilon\leq \sum_{j=1}^m w_{j}f_{jq}(\xx_i)+M\cdot b(i,q).\\
\end{align*}

We define 
\begin{align*}
B(i):&=\ind \Big\{\argmax_{k\in[K]}\sum_{j=1}^m w_{j}f_{jk}(\xx_i) = y_i\Big\}
\\
&=\land_{q\in[K]}\:b(i,q) , \:\forall i\in \mathcal{I},
\end{align*}
    
which can be represented using the following constraints in the MIO formulation:

\begin{align*}
    &B(i)\in\{0,1\},\\ 
    &B(i)\leq b(i,q),\:\forall q\in [K],\\
    &B(i)\geq \sum_{q\in[K]} b(i,q)-(k-1).
\end{align*}

Combining all components, we obtain the final complete MIO formulation:

\begin{subequations} \label{eqn: vanilla multi}
\begin{align}
\max_{\mathbf{w}} \hspace{0.2cm} &\sum_{i\in\Ic}B(i)\\
s.t. \hspace{0.2cm}& \mathbf{w}\in \R^m_{\geq 0},\\ &\sum_{j=1}^m w_j=1,\\
&b(i,q)\in\{0,1\},\:\forall i\in \mathcal{I},q\in[K],\\ 
&M\cdot (1-b(i,q))+\sum_{j=1}^m w_{j}f_{jy_i}(\xx_i)\geq \sum_{j=1}^m w_{j}f_{jq}(\xx_i),\:\forall i\in \mathcal{I},q\in[K],
\\
&\sum_{j=1}^m w_{j}f_{jy_i}(\xx_i)+\epsilon\leq \sum_{j=1}^m w_{j}f_{jq}(\xx_i)+M\cdot b(i,q),\:\forall i\in \mathcal{I},q\in[K],
\\
&B(i)\in\{0,1\},\:\forall i\in \mathcal{I},\\ 
       &B(i)\leq b(i,q),\:\forall q\in [K],\: i\in \mathcal{I},\\
       &B(i)\geq \sum_{q\in[K]} b(i,q)-(k-1),\:\forall i\in \mathcal{I},
\end{align}
\end{subequations} 
where $M=2$ is big enough and $\epsilon>0$ to be small.

\section{More Detailed and Complete Pipeline}\label{appendix C}
\subsubsection*{Step 1: split datasets}
For a given dataset used in binary or multi-class classification, we partition it into four disjoint subsets: \texttt{data\_single}, \texttt{data\_opt}, \texttt{data\_val}, and \texttt{data\_test}, as outlined in \Cref{table:data division}.

\subsubsection*{Step 2: train $m$ base learners}

Each base learner is a CART model trained on randomly sampled subsets of data instances and features from \texttt{data\_single}, similar to the approach used in Random Forests. This process ensures that the individual CARTs exhibit distinct predictive strengths and varying levels of accuracy and specialization across different regions of the feature space. Such diversity enhances the effectiveness of ensemble methods and justifies the application of non-uniform weighting for each data instance according to its specific features.

The default down-sampling ratio for both data instances and features is uniformly sampled from the range 
$[0.5,0.9]$. A higher down-sample ratio allows for larger training datasets, leading to better performance of individual models. However, it also increases the likelihood of overlapping training data, which can reduce the differentiation between individual base learners.

\subsubsection*{Step 3: first iteration}
During the first training iteration, we perform two tasks: (a) initialize the set of potential ensemble weights $\mathbb{W}$ according to \Cref{appendix A: weight init}, and (b) train a preliminary 
OPT over \texttt{data\_opt} using Interpretable AI's package OptimalTreePolicyMaximizer (\url{https://docs.interpretable.ai/stable/OptimalTrees/reference/#IAI.OptimalTreePolicyMaximizer}) with various configurations discussed in \Cref{subsec: different configurations}. We then calculate the AUC over \texttt{data\_val} to select the best configuration, which will remain fixed for subsequent iterations.

\subsubsection*{Step 4: adapt over iterations}

This process is repeated for up to 10 iterations or until the OPT structure stabilizes, meaning it remains unchanged even if Step 4 is repeated.

During this step, (a) new candidates for $\mathbb{W}$ are adaptively generated based on the current OPT's structure, the existing $\mathbb{W}$, and the dataset; (b) the best $k$ candidates are selected, if necessary, to prevent $\mathbb{W}$ from growing excessively or increasing the complexity of the Adaptive Forests model; and (c) a new reward matrix is computed, and the OPT is re-trained accordingly.

\paragraph{Generate new candidates for $\mathbb{W}$:} Using MIO algorithms detailed in \ref{subsec: MIO}, we propose new candidates for $\mathbb{W}$ that are promising at either the leaf level or the tree level, designed for exploration or exploitation. This weight generation strategy dynamically adapts to the current OPT structure and the existing $\mathbb{W}$, while also adjusting across different datasets to achieve improved performance.

\paragraph{Pick the best $k$ candidates and update $\mathbb{W}$:} From the newly generated candidates, an optional step can be performed to select the top $k$ most promising candidates to be added to $\mathbb{W}$, where $k$ is a configurable hyperparameter and ``promising" is defined by the objective in MIO \ref{refer error}. This step helps to manage the growth of $\mathbb{W}$, reduce the complexity of the Adaptive Forests model, prevent overfitting, and optimize computation time. The $\mathbb{W}$ set is then updated by removing unused candidates, adding new ones, and reintroducing previously used candidates, provided they were selected by OPT in at least one historical iteration.

\paragraph{Update reward matrix and re-train OPT:}
With the changes in $\mathbb{W}$, a new reward matrix must be computed, followed by training a new OPT.

\subsubsection*{Step 5: report final performance over test data}

In the end, we obtain the iteratively refined version of our Adaptive Forests model. The final performance is evaluated using the metric appropriate for each specific problem over \texttt{data\_test}. 

\subsubsection*{Code Availability}

\begin{center}
\url{https://github.com/YubingCui713/Adaptive-Forests-For-Classification}
\end{center}

\end{document}